\documentclass[11pt]{article}
\usepackage[a4paper,margin=1in]{geometry}
\usepackage{amsmath,amssymb,amsfonts}
\usepackage{graphicx}
\usepackage{booktabs}
\usepackage[hidelinks]{hyperref}
\usepackage{authblk}
\usepackage[sc]{titlesec}
\usepackage{caption}
\usepackage{microtype}
\title{\bfseries Ablation-Corrected Evaluation of Attribution Maps\\
in Echocardiographic Ejection-Fraction Models}

\author[1]{Hyunkyung Han}
\author[1,2]{Min Jung Kim}
\affil[1]{Department of Integrative Medicine, Yonsei University College of
Medicine, Seoul, Republic of Korea}
\affil[2]{Department of Radiology, Research Institute of Radiologic Science,
Yonsei University College of Medicine, Seoul, Republic of Korea}
\date{}

\begin{document}
\maketitle

\begin{abstract}
Attribution maps for echocardiographic ejection-fraction models are evaluated by
their overlap with an expert left-ventricular annotation, compared against a
chance level that is computed from an area ratio rather than measured. We measure
it. Two architectures trained on the same task attain overlap at 3.55 and 4.20
times measured chance, an eighteen percent difference a reader would take as the
size of the gap between them. It is not. Replacing the annotated ventricle with a
composition-matched surrogate changes one model's prediction six times more than
replacing an equal-area control region and the other's twice, a factor of three;
on sixty-two percent of cases for the second, neither the ventricle nor the
attributed region moves the prediction at all. The reliability of this
measurement, from two independent intervention batches, is 0.90 to 0.98. Overlap
does not track the difference: scored against an ablation-derived label of which
cases the model depends on the ventricle for, it reaches an area under the curve
of 0.52 and 0.67. The design of the ablation also determines its answer, since
ablating only the two annotated frames leaves the ventricle indistinguishable
from a control region while ablating across all frames does not. A pediatric
cohort reproduces all of these. We give the measured chance level, the
symmetric-ablation protocol, the reliability estimate, and the
ablation-prediction score as things to report alongside overlap.
\end{abstract}

\noindent\textbf{Keywords:} attribution methods; echocardiography; ejection
fraction; model interpretability; evaluation protocols; ablation.

\vspace{1em}

\section{Introduction}
\label{sec:intro}

An attribution map for a deep model that estimates ejection
fraction from echocardiography is judged by where it points. The usual test is
overlap: the fraction of the attributed region that falls inside an expert
tracing of the left ventricle, reported as a ratio to the value that overlap
would take by chance. A value several times chance is read as evidence that the
model attends to the anatomy a clinician would attend to. That reading is what
makes the score worth reporting, and it is the reading we examine here.

The chance level is ordinarily computed rather than measured: the annotated
region occupies some fraction of the frame, and that fraction is taken as the
value an uninformative map would attain. It is not. Constructing random maps and
applying the identical scoring procedure gives a different value, and the gap
between the two is not constant across models or cohorts.

More consequentially, the corrected score does not distinguish models that differ
in what they use. Two architectures trained on the same data---a video
masked-autoencoder and an R(2+1)D convolutional network---attain overlap at 3.55
and 4.20 times measured chance, a difference of eighteen percent. Ablating the
ventricle shows that one of them depends on it six times more than on an
equal-area control region and the other twice, a difference of a factor of three;
on 62\% of cases for the second, neither the ventricle nor the region the
attribution itself selected moves the prediction at all.

The reason overlap cannot make this distinction is structural. It compares an
attribution against an annotation, and neither term involves the model
(Fig.~\ref{fig:concept}a). Whether the model uses the annotated region is a third
quantity, and it must be measured separately: by replacing that region with a
composition-matched surrogate and recording how the prediction moves, against the
movement produced by replacing an equal-area region elsewhere. That measurement
is what the overlap score omits, and supplying it changes what the score is taken
to mean.

This paper contributes the following.

\begin{itemize}
\item A measured rather than computed chance level for attribution overlap in
echocardiography, obtained by scoring random maps through the identical pipeline
(Section~\ref{sec:chance}). It is 0.164 on the adult cohort and 0.124 on the
pediatric one; an area ratio does not show that difference.

\item A grid-level ablation with a composition-matched surrogate and an
equal-area control, separating the model's dependence on the ventricle from the
effect of substitution itself (Section~\ref{sec:ablation}). Ablating the
attribution's own top region under the same control forecloses the objection that
the model uses something the annotation omits (Section~\ref{sec:symmetric}).

\item A reliability estimate for that ablation, from two fully independent
intervention batches, of 0.90 to 0.98 (Section~\ref{sec:reliability}),
establishing that the case-level classification is not an artefact of measurement
noise.

\item The finding that the ablation's design determines its answer: ablating only
the two annotated frames leaves the ventricle indistinguishable from a control
region, while ablating across all frames does not (Section~\ref{sec:design}).

\item An ablation-derived score to report alongside overlap, which separates the
two architectures where overlap does not, and which varies eight times less
across training seeds (Sections~\ref{sec:auc} and \ref{sec:seeds}).
\end{itemize}

\section{Related Work}
\label{sec:related}

\subsection{Automated EF estimation}
Automated cardiac function assessment from echocardiography was established at
scale by EchoNet-Dynamic \cite{ouyang2020}, which trained spatiotemporal
convolutional networks on beat-level clips and matched expert variability in EF
estimation; the paradigm has been extended to pediatric cohorts \cite{reddy2023}
and to open segmentation benchmarks \cite{leclerc2019}. Architecturally, EF
estimators have progressed from factorized 3D convolutions \cite{tran2018} to
video transformers with self-supervised pretraining \cite{tong2022}. We evaluate
one representative of each family, and Section~\ref{sec:results} shows that the
choice matters more than the attribution literature has assumed.

\subsection{Attribution and its evaluation}
Post-hoc attribution scores input importance to explain a prediction: Grad-CAM
\cite{selvaraju2017} and Integrated Gradients \cite{sundararajan2017} for
convolutional networks; for transformers, raw attention is an unreliable
explanation \cite{jain2019,wiegreffe2019}, motivating relevance-propagation
methods that combine attention with gradients \cite{chefer2021}. We adopt Chefer
relevance for VideoMAE and Grad-CAM for R(2+1)D.

Whether such a map is faithful is assessed by perturbation---deletion and
insertion curves \cite{petsiuk2018}, remove-and-retrain \cite{hooker2019}, and
model-randomization sanity checks \cite{adebayo2018}---several of which show that
popular methods fail basic tests. These procedures share a structure: they rank
input positions \emph{by the attribution itself} and measure how the prediction
degrades. They therefore presume the ranking is meaningful in order to evaluate
it. Ablating an annotated region instead supplies a reference point that no
attribution enters, which is what makes the comparison in
Section~\ref{sec:auc} possible.

\subsection{Where the failure is located}
Koo and Eddy \cite{koo2019} established that a model can attain high accuracy
while its attributions fail to match a known feature, and located the cause in
the architecture: filter size and pooling determine whether first-layer filters
learn whole features or partial ones. A complementary route pursues intrinsic
interpretability for EF regression \cite{ghamary2025}, motivated by the concern
that post-hoc explanations do not shape a model's internal reasoning. Our results
locate a third possibility: the explanation may be faithful and the architecture
adequate while the evaluation metric fails to distinguish that case from its
opposite.

\subsection{Chance correction in overlap scores}
The problem of an overlap statistic that is not zero under independence has an
established solution---subtracting the expectation and rescaling, as in the
adjusted Rand index \cite{hubert1985}. In genomic colocalisation, Salvatore
\textit{et al.} \cite{salvatore2020} argue against the raw Jaccard index for the
same reason, and Chikina and Troyanskaya \cite{chikina2012} compute an exact
combinatorial null for interval proximity. In medical imaging the corresponding
correction is not standard: overlap between an attribution and an annotation is
reported against an area ratio when it is reported against anything at all.

\begin{figure}[!t]
\centering
\includegraphics[width=\linewidth]{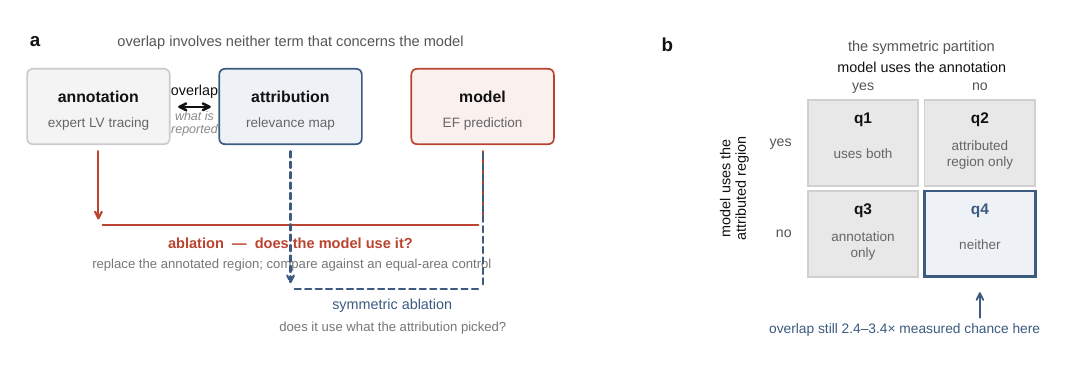}
\caption{The overlap score and what it omits. (a) Overlap compares an attribution
against an annotation; neither term involves the model. Whether the model uses
the annotated region is measured by ablation, and whether it uses the region the
attribution selected by a symmetric ablation under the same control. (b) The
resulting four-way partition. All statements about cases where the model does not
use the ventricle are restricted to the fourth cell, where neither intervention
moves the prediction.}
\label{fig:concept}
\end{figure}

\section{Methods}
\label{sec:methods}

\subsection{Measured chance level}
\label{sec:chance}
The overlap score is the fraction of the thresholded attribution that falls
inside the annotated ventricle. Its chance value is ordinarily taken to be the
area fraction of the annotation, on the reasoning that an uninformative map
distributes its mass uniformly. Attribution maps are not uniform: they have
spatial structure imposed by the architecture---a token grid in the transformer,
a downsampled feature map in the convolutional network---and that structure is
present whether or not the map carries information about the input.

We therefore measure the chance level rather than compute it. For each case we
generate random maps with the same support and the same thresholding rule, score
them through the identical pipeline, and take the mean. The annotation is the
expert tracing rasterised to the model's $7\times7$ grid; a cell counts as
annotated when more than 20\% of its area falls inside the tracing. The value is
not a constant of the task: it is 0.164 on the adult cohort and 0.124 on the
pediatric one, because a pediatric ventricle occupies a smaller share of the
grid. Reporting raw overlap for the two side by side would read that difference
as a difference in the models.

\subsection{Grid-level ablation}
\label{sec:ablation}
Whether the model uses the annotated region is measured by removing it. Removal
must not itself carry information: setting pixels to zero produces a black region
in an ultrasound sector, which is a signal in its own right and one the model may
respond to independently of what was removed.

We therefore replace rather than zero. For each case we draw a pixel pool from the
region inside the ultrasound sector and outside the annotation, and fill the
annotated cells by sampling from it. The replacement preserves local intensity
statistics while destroying the structure. The same procedure applied to an
equal-area region elsewhere gives the control, so that
\begin{equation}
u = \big| \mathrm{EF}_{\mathrm{abl}} - \mathrm{EF}_{0} \big| -
    \big| \mathrm{EF}_{\mathrm{ctl}} - \mathrm{EF}_{0} \big|
\label{eq:usage}
\end{equation}
isolates the effect of removing the ventricle from the effect of substitution as
such. Ablation is applied to all frames of the clip; Section~\ref{sec:design}
shows why.

\subsection{Symmetric ablation}
\label{sec:symmetric}
That the model does not use the annotated ventricle leaves open that it uses
something else which the attribution correctly identifies---that the annotation,
not the attribution, is at fault. We therefore ablate the attribution's own
selected region under the same control, restricted to cells outside the
annotation so that the two interventions do not overlap.

This partitions cases four ways (Fig.~\ref{fig:concept}b). The cell in which the
model uses the attributed region but not the annotation is the one the objection
describes, and we report its size. All statements about cases where the model does
not use the ventricle are restricted to the cell where neither intervention moves
the prediction.

\subsection{Reliability of the intervention}
\label{sec:reliability}
The four-way classification thresholds a quantity measured with noise: the
replacement is drawn at random, so $u$ is one draw from an intervention
distribution rather than a fixed property of the case. A threshold applied to an
unreliable measurement selects a contaminated subset.

We therefore estimate the reliability directly. Two fully independent batches of
three draws each are run per case, with the pixel pool, the control position and
the random stream regenerated for every draw. Under a parallel-replicate model
the covariance between batch means estimates the between-case variance of the
latent quantity, so
\begin{equation}
\lambda = \frac{\operatorname{Cov}(A,B)}{\operatorname{Var}\!\big((A+B)/2\big)}
\label{eq:lambda}
\end{equation}
is the reliability of the pooled measurement.

\subsection{Ablation-prediction score}
\label{sec:auc}
Given the ablation as a case-level label, the question of whether an attribution
is informative becomes a ranking question: does a higher overlap identify a case
where the model depends more on the ventricle? We take
\begin{equation}
\mathrm{AUC}\big(\mathbf{1}[u > \tau],\ \mathrm{overlap}\big)
\label{eq:auc}
\end{equation}
with $\tau$ at a fixed quantile of $u$. Unlike the overlap itself, this quantity
involves the model, and it is the quantity we recommend reporting.

\subsection{Data, models and attributions}
\label{sec:setup}
\textit{Data.} EchoNet-Dynamic \cite{ouyang2020} provides 10{,}030 apical
four-chamber studies with expert left-ventricular tracings at end-systole and
end-diastole; we use the official test split (1{,}276 studies) throughout.
EchoNet-Pediatric \cite{reddy2023} provides the external cohort, restricted to
files with exactly two annotated frames (1{,}182 studies). Clips are sampled at
32 frames with a period of 2, starting from a position chosen to cover both
annotated frames, and resized to $112\times112$.

\textit{Models.} Both models regress EF through a sigmoid output. R(2+1)D
\cite{tran2018} is initialised from Kinetics pretraining; VideoMAE
\cite{tong2022} from self-supervised pretraining on the same echocardiographic
corpus. Both are fine-tuned for 50 epochs with AdamW, learning rate
$1\times10^{-4}$, weight decay 0.05, batch size 32 and a one-cycle schedule with
5\% warmup. The best validation checkpoint is used: 4.69 mL MAE for R(2+1)D
(epoch 45) and 6.82 for VideoMAE (epoch 43). Two further VideoMAE models were
trained under identical settings with different random seeds, giving 6.94 and
6.94.

\textit{Attributions.} Chefer relevance \cite{chefer2021} is computed on
VideoMAE by propagating gradient-weighted attention through the encoder blocks
and reshaping to the $(T,7,7)$ token grid. Grad-CAM \cite{selvaraju2017} is
computed on R(2+1)D at the final residual stage, whose output is
$(512,4,7,7)$. Both therefore live on a $7\times7$ spatial grid, which is the
grid on which the annotation is rasterised and the ablation applied. Maps are
min--max normalised per frame and thresholded at the 70th percentile.

\textit{Ablation.} Each ablated cell is filled by sampling, with replacement,
from the pixels inside the ultrasound sector and outside the annotation, defined
by an intensity threshold of 10 on the 0--255 scale. The control comprises the
same number of cells drawn uniformly from those belonging to neither the
annotation nor the attribution's selected region. Reliability uses two
independent batches of three draws each, all stochastic elements regenerated per
draw.

\textit{Reproducibility.} Random seeds are fixed for \texttt{random},
\texttt{numpy}, \texttt{torch} and \texttt{torch.cuda}. Code and per-case
outputs are available on publication.

\section{Results}
\label{sec:results}

\subsection{Overlap puts the two architectures 18\% apart}
\label{sec:overlap}
Both models attain overlap well above the measured chance level: $3.55\times$ for
VideoMAE, averaged over three seeds, and $4.20\times$ for R(2+1)D. On the overlap
alone the second is 18\% higher---three times the seed-to-seed variation of 5.7\%
reported in Section~\ref{sec:seeds}---and a reader would take that as a real
difference in degree.

The two models are not equally accurate: R(2+1)D reaches 4.69 mL validation MAE
against 6.82 for VideoMAE. A reader comparing them would therefore expect the
better model to be the better grounded one, and the overlap agrees, by 18\%.

\subsection{Ablation puts them a factor of three apart}
\label{sec:sep}
\begin{figure}[!t]
\centering
\includegraphics[width=\linewidth]{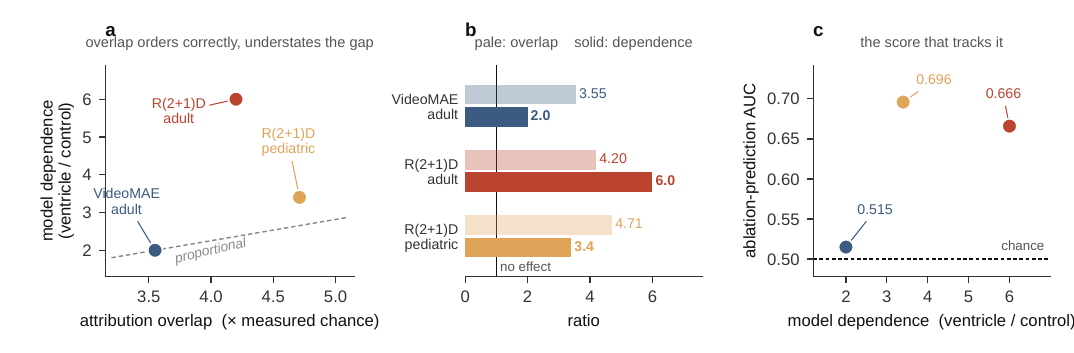}
\caption{Attribution overlap orders the three settings correctly but understates
the gap between them. (a) Overlap against ablation-measured dependence; the
dashed line marks where the points would fall if overlap were proportional to
dependence. (b) The same values as ratios; pale bars are overlap, solid bars
dependence, and the vertical line marks an intervention with no effect. (c) The
ablation-prediction AUC rises with dependence, from chance for VideoMAE to
0.67--0.70 for R(2+1)D.}
\label{fig:main}
\end{figure}

Replacing the annotated ventricle changes the two models very differently
(Table~\ref{tab:main}, Fig.~\ref{fig:main}). The R(2+1)D prediction moves six
times more than when an equal-area control region is replaced; the VideoMAE
prediction moves twice as much.

The gap is larger at the case level. Under the symmetric partition of
Section~\ref{sec:symmetric}, 62\% of VideoMAE cases fall in the cell where
neither the ventricle nor the attributed region moves the prediction, against
13\% for R(2+1)D. The overlap put the two 18\% apart; the ablation puts them a
factor of three apart. The overlap orders them correctly and understates the gap
by an order of magnitude.

The cell corresponding to the incomplete-annotation objection---the model uses the
attributed region but not the ventricle---contains 20 of 600 VideoMAE cases
(3.3\%) and 60 of 1255 R(2+1)D cases (4.8\%). The objection describes a real but
small minority.

Three quantities order the two models the same way: accuracy (4.69 against 6.82
mL), dependence on the annotated ventricle ($6.0\times$ against $2.0\times$), and
the ablation-prediction score of Section~\ref{sec:pred} (0.67 against 0.52). The
overlap orders them the same way too, but reports the gap as 18\% where the other
three report a factor of three. It is not that the overlap is uninformative; it
is that its magnitude cannot be read as the magnitude of anything the reader
cares about.

\begin{table}[!t]
\caption{Overlap, dependence and the proposed score across three settings.
$|\Delta|$ values are on each model's own EF scale (0--1 for VideoMAE, 0--100 for
R(2+1)D); compare the ratios.}
\label{tab:main}
\centering
\setlength{\tabcolsep}{3.2pt}
\small
\begin{tabular}{lccccc}
\toprule
 & overlap & dependence & $q_4$ & $q_4$ overlap & AUC \\
 & ($\times$ chance) & (LV / ctl) & & ($\times$ chance) & \\
\midrule
VideoMAE, adult    & 3.55 & \textbf{2.0} & 62\% & 3.39 & 0.515 \\
R(2+1)D, adult     & 4.20 & \textbf{6.0} & 13\% & 2.42 & 0.666 \\
R(2+1)D, pediatric & 4.71 & \textbf{3.4} & 17\% & 2.42 & 0.696 \\
\bottomrule
\end{tabular}
\end{table}

\subsection{The measurement is reliable}
\label{sec:lam}
The four-way classification rests on a threshold applied to a noisy quantity, so
its reliability determines whether the subsets are contaminated. Across the three
ablation conditions the pooled reliability is 0.899, 0.957 and 0.981, with
between-batch correlation 0.92 to 0.96. At these values the attenuation of a
threshold-selected contrast is under 3\%, and the classification is not an
artefact of measurement noise.

\subsection{Overlap does not predict model dependence}
\label{sec:pred}
Scored against an ablation-derived label of which cases the model depends on the
ventricle for, the overlap achieves an AUC of 0.515 for VideoMAE and 0.666 for
R(2+1)D, stable across thresholds from the 50th to the 80th percentile of $u$.

The first value is chance. For VideoMAE, knowing that a case has high overlap
conveys nothing about whether the model used the ventricle in that case. The
second is informative but modest: even where the model depends strongly on the
annotated anatomy, the overlap ranks cases only somewhat better than a coin. The
two AUCs differ by 0.15 while the overlap scores differ by $0.65\times$ chance.
The quantity that separates the two architectures is the one that involves the
model.

\subsection{The design of the ablation determines its answer}
\label{sec:design}
\begin{figure}[!t]
\centering
\includegraphics[width=\linewidth]{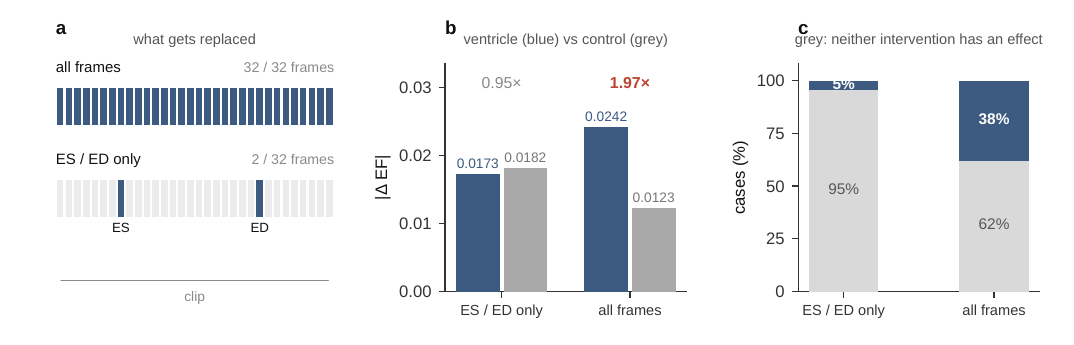}
\caption{The design of the ablation determines its answer. (a) The two
protocols. (b) Ablating only the annotated frames leaves the ventricle
indistinguishable from a control region; ablating across the clip does not.
(c) The share of cases in which neither intervention moves the prediction falls
from 95\% to 62\%.}
\label{fig:design}
\end{figure}

Restricting the ablation to the two frames carrying an expert tracing---the
natural choice, since that is where the annotation is defined---gives a different
conclusion (Fig.~\ref{fig:design}). Ablating only the annotated frames leaves the
ventricle indistinguishable from a control region ($0.0173$ against $0.0182$,
a ratio of $0.95$), and 95\% of cases fall in the cell where nothing moves the
prediction. A study stopping there would report that the model does not use the
ventricle at all---a conclusion the same measurement contradicts when the region
is ablated across the clip.

The reason is that ejection fraction is estimated from the whole sequence.
Removing the ventricle from two of 32 frames leaves 30 intact, and the model
recovers. This is a property of the protocol rather than of the model, and it is
not visible unless both protocols are run.

\subsection{Seed variation}
\label{sec:seeds}
\begin{figure}[!t]
\centering
\includegraphics[width=\linewidth]{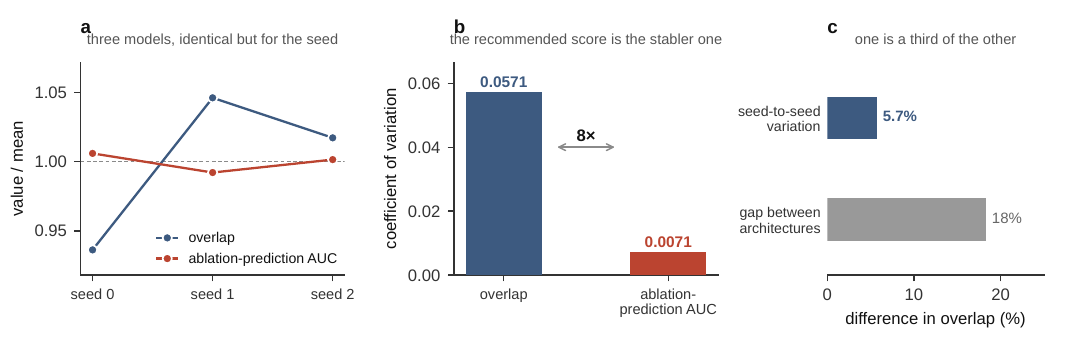}
\caption{Seed variation. (a) Values relative to their mean across three models
identical but for the random seed. (b) The overlap varies eight times more than
the ablation-prediction AUC. (c) That variation is a third of the difference the
overlap reports between the two architectures.}
\label{fig:seeds}
\end{figure}

Three independently trained VideoMAE models, differing only in random seed, give
validation MAE of 6.82, 6.94 and 6.94. The overlap across them is 0.4649, 0.5196
and 0.5052, a coefficient of variation of 0.0571; the ablation-prediction AUC is
0.5153, 0.5082 and 0.5130, a coefficient of variation of 0.0070
(Fig.~\ref{fig:seeds}).

The quantity we recommend reporting is not only the more informative of the two
but the more stable, by a factor of eight. The overlap moves 5.7\% while the model
does not materially change---a third of the difference that separated the two
architectures in Section~\ref{sec:overlap}, which is a reason to read that
difference with caution.

\subsection{The temporal axis carries no signal}
\label{sec:temporal}
Ejection fraction is defined by two frames: end-diastole and end-systole. A model
estimating it from a 32-frame clip must locate them, and an attribution that
reflected that would concentrate its mass there.

It does not. Measured as the ratio of attribution mass in the tubelets containing
the two annotated frames to the mass expected under uniform distribution over the
clip, the concentration is 0.985---indistinguishable from uniform. The spatial and
temporal axes therefore behave differently in the same map: the attribution
locates the ventricle at several times chance and locates the two frames that
define the quantity being predicted at chance exactly.

This is the same failure mode as Section~\ref{sec:design}, seen from the other
side. There, ablating two frames of 32 left the model unaffected because the
remaining 30 sufficed; here, the attribution spreads across all 32 because no
small set of frames is individually necessary. Both observations say that the
temporal structure of the prediction is distributed.

\subsection{External validation}
\label{sec:external}
Running the same protocol on a pediatric cohort reproduces the adult picture in
every respect that matters to the argument.

The measured chance level is 0.124, below the adult 0.164, because a pediatric
ventricle occupies a smaller share of the grid. Overlap is $4.71\times$ that
level, above the adult $4.20\times$. Read on the overlap alone, the model looks
better grounded on children than on adults.

The ablation says otherwise, and says the same thing in both cohorts. Replacing
the ventricle moves the prediction 3.4 times more than replacing a control region
(6.0 in adults); the intervention reliability is 0.968 (0.981); and on the 16.5\%
of cases where neither intervention moves the prediction, overlap still stands at
$2.42\times$ measured chance---the adult value to two decimal places. The
ablation-prediction AUC is 0.696 against the adult 0.666.

The divergence between where the attribution points and what the model uses is
therefore not a property of the adult data, of the particular fine-tuning run, or
of the patient population.

\section{Discussion}
\label{sec:discussion}

\subsection{What the overlap measures}
The overlap score compares two objects, an attribution and an annotation, neither
of which involves the model. It is therefore a statement about where the
annotated anatomy is and where the attribution points, and it remains that
statement whether or not the model uses the anatomy. Our two architectures make
this concrete: they overlap the ventricle at rates 18\% apart while differing
threefold in how much of the prediction the ventricle accounts for.

This is not a claim that the score is meaningless. An attribution that pointed at
the sector boundary would score at chance, and that would be informative. The
claim is narrower: above chance, the score orders models correctly but does not
report the size of the difference between them.

\subsection{Why the chance level must be measured}
Computing the chance level from an area ratio assumes an uninformative
attribution is spatially uniform. It is not: the transformer's relevance lives on
a token grid, the convolutional model's on a downsampled feature map, and both
have structure that survives when the map carries no information about the
particular input. A random map scored through the same pipeline therefore does
not attain the area fraction, and the discrepancy differs between architectures
and between cohorts---which are precisely the comparisons a reader is likely to
make.

\subsection{The ablation is a design choice, not a measurement}
Section~\ref{sec:design} is the result we would most want a reader to carry away.
The same intervention, applied to the two annotated frames or to the whole clip,
gives opposite conclusions about whether the model uses the ventricle. Neither is
wrong as an experiment; they answer different questions. But only one of them
answers the question the overlap score is being compared against, and a paper
reporting the two-frame version would state, in good faith, a conclusion its own
data contradicts.

The same applies to the substitution. Filling the ablated region with zeros
produces a black region in an ultrasound sector---an artefact the model can
respond to independently of what was removed, and one that biases the measurement
in an unknown direction. Composition-matched replacement removes the structure
while preserving the local statistics, which is the intervention the question
requires.

\subsection{Relation to attribution evaluation elsewhere}
The structure of the problem is not specific to echocardiography. Any evaluation
that scores an attribution against an annotation compares two model-independent
objects, and the same three-way confusion arises: whether the feature is
findable, whether the model uses it, and whether the attribution points at it.
The pointing game in vision, rationale overlap in natural language processing,
and motif recovery in genomic sequence models share it.

Deletion and insertion curves are sometimes offered as the model-side check. They
are not one: those curves rank positions by the attribution itself, so they
presume the ranking is meaningful in order to evaluate it. Ablating the annotated
region instead supplies a reference point that no attribution enters, and it is
that independence which makes the comparison in Section~\ref{sec:pred} possible.

\subsection{Limitations}
Our ablation is an input--output measurement. It establishes that replacing the
ventricle does not change the prediction, which is weaker than establishing that
the model's computation does not involve it; a model routing the equivalent
information through the surrounding myocardium would be scored as not using the
annotated region. We use the symmetric ablation of Section~\ref{sec:symmetric} to
bound one version of this concern, but not all of them.

The composition-matched replacement preserves local intensity statistics and
destroys structure, so what we call dependence on the ventricle is dependence on
its structure rather than on its position or brightness. A model relying only on
where the dark region sits would be scored as not using it.

Two architectures do not establish a range. We report that they differ, not how
far apart architectures can be. Seed repetition was performed for one of the two.

\section{What to Report}
\label{sec:report}

We suggest four things be stated alongside an attribution overlap.

\emph{The chance level, measured.} Random maps scored through the same pipeline,
not an area ratio. Section~\ref{sec:external} shows it differs between cohorts by
a third.

\emph{The ablation and its control.} The prediction change when the annotated
region is replaced with a composition-matched surrogate, against the change when
an equal-area region is, stated with the number of frames ablated.

\emph{The reliability of that ablation.} Two independent intervention batches and
the resulting $\lambda$. A threshold applied to an unreliable measurement selects
a contaminated subset, and $\lambda$ is what tells a reader whether it did.

\emph{The ablation-prediction AUC.} Whether the overlap identifies the cases where
the model depends on the annotation. This is the quantity that separated our two
architectures where the overlap did not, and the one that varies least across
training seeds.

None of these requires an experiment beyond the ones already performed to produce
the attribution: the chance level is a scoring pass over random maps, and the
ablation is fewer forward passes than the attribution itself.

\end{document}